%% file: egpaper_for_review.tex
\documentclass[10pt,twocolumn,letterpaper]{article}

\usepackage{wacv}
\usepackage{times}
\usepackage{epsfig}
\usepackage{graphicx}
\usepackage{amsmath}
\usepackage{bm}
\usepackage{amssymb}
\usepackage{paralist}
\newcommand{\vz}{\mathbf{z}}
\newcommand{\vmu}{\bm{\mu}}
\newcommand{\vx}{\mathbf{x}}

\newcommand{\vd}{{\bm{\delta}}}
\newcommand{\vphi}{{\bm{\phi}}}

\newcommand{\vv}{\mathbf{v}}

\newcommand{\vI}{\mathbf{I}}
\newcommand{\vzero}{\bf{0}}

\newcommand{\expect}{\mathbb{E}}

\newcommand{\Normal}{\mathcal{N}}

\usepackage{hyperref}

\wacvfinalcopy 

\def\wacvPaperID{203} 

\ifwacvfinal\fi
\begin{document}

\title{Jointly Trained Image and Video Generation using Residual Vectors}
\author{ Yatin Dandi \hspace{1cm} Aniket Das \hspace{1cm} Soumye Singhal \hspace{1cm} Vinay P. Namboodiri \hspace{1cm} Piyush Rai \\
Indian Institute of Technology, Kanpur\\
{\tt\small \{yatind,aniketd\}@iitk.ac.in,singhalsoumye@gmail,\{vinaypn,piyush\}@cse.iitk.ac.in }
}
\maketitle
\ifwacvfinal\thispagestyle{plain}\fi

\begin{abstract}
     In this work, we propose a modeling technique for jointly training image and video generation models by simultaneously learning to map latent variables with a fixed prior onto real images and interpolate over images to generate videos. The proposed approach models the variations in representations using residual vectors encoding the change at each time step over a summary vector for the entire video. We utilize the technique to jointly train an image generation model with a fixed prior along with a video generation model lacking constraints such as disentanglement. The joint training enables the image generator to exploit temporal information while the video generation model learns to flexibly share information across frames. Moreover, experimental results verify our approach's compatibility with pre-training on videos or images and training on datasets containing a mixture of both. A comprehensive set of quantitative and qualitative evaluations reveal the improvements in sample quality and diversity over both video generation and image generation baselines.  We further demonstrate the technique's capabilities of exploiting similarity in features across frames by applying it to a model based on decomposing the video into motion and content. The proposed model allows minor variations in content across frames while maintaining the temporal dependence through latent vectors encoding the pose or motion features.   
\end{abstract}

\section{Introduction}
The success of deep generative models in recent years has been most visible and profound in the task of image generation. Two of the prominent approaches have been Generative Adversarial Networks ~\cite{GAN} and Variational Auto-Encoders ~\cite{VAE}. These models involve mapping low-dimensional latent variables to images using a Convolutional Neural Network. In image generation models, a single latent variable for an image with an Isotropic Gaussian prior is often sufficient to capture the underlying factors of variation. Generation and unsupervised representation learning of videos, however, poses additional difficulties due to the inherently high stochasticity and the need to capture temporal dependence across frames. This suggests that modeling of the generation (and inference model for the case of VAEs) can play a crucial role in determining the model's performance. Moreover, careful design of the prior can lead to representations with specific interpretations and uses in downstream tasks.

A common approach has been to model the process as a sequential generation of the image representations for the individual frames. In such models, the temporal dependence across the latent variables for individual frames is often captured through an RNN ~\cite{SVG,SAVP,MOCOGAN,DSA,VIDEOFLOW,IMPROVEDVRNN,MULTISCALE,MCNET,DRNET}. While this approach provides the flexibility to generate videos, it
prohibits direct access to the distribution of images in latent space as the generation of the images in their full extent of variation must proceed through the computationally expensive generation of entire videos. This may be avoided in datasets where any arbitrary image corresponds to the first frame of a video, but in most cases, attributes such as actions are expressed in the later frames.  We propose an approach to learn video generation while simultaneously imposing a fixed chosen prior on the image latent vectors. The proposed approach confers the following advantages over other baselines such as MoCoGAN~\cite{MOCOGAN}:
\begin{enumerate}
    \item Training the model on a set of videos results in an image generator having superior sample quality than a baseline generator trained only on images due to the ability to exploit temporal information to improve sample quality during the learning process. The image generator doesn't require the computational overhead of the video generation process.
    \item The approach allows the model to retain the advantages provided by an image-only generator and directly borrow modifications and downstream uses from the large body of work in image generation relying on a fixed, known prior \cite{BETAVAE,INFOGAN,STYLEGAN}.
    \item The ability to enforce a fixed prior in the image latent space makes the proposed video generation models compatible with pre-training on images or videos as well as training on datasets containing a mixture of both. This compatibility is especially important in the present times, considering the relatively large abundance of image data compared to videos.
    \item Interpolation in video space can be performed using the standard techniques for interpolation in image latent space. (Figure 10 and 11 in Appendix)
\end{enumerate}

\begin{figure}
  \centering
  \includegraphics[width=\linewidth]{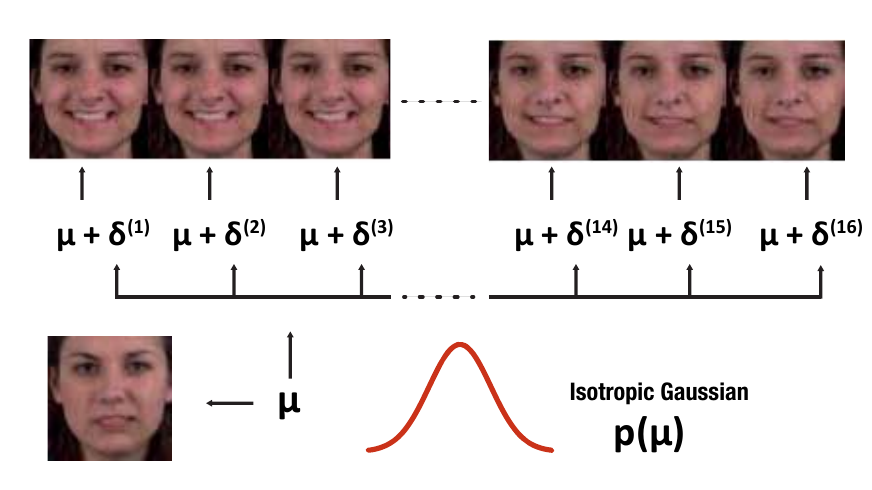}
  \caption[]{The proposed approach for joint image and video generation. The first step of the video generation process involves sampling a summary vector $\vmu$ which encodes an image capturing the shared factors of variation across frames. Subsequently, an RNN is used to generate the entire video by interpolating over $\vmu$ in the image latent space.}
  \label{fig:intro_image}
\end{figure}

Apart from the dependence of the current frame on the previous frames, another common property of most natural and synthetic video datasets is the presence of the same set of objects across all the frames within a video which undergo motion following certain rules of dynamics. Thus, in a dataset of human action videos, variations within a video are limited to changes in the person's pose while his overall appearance remains nearly constant, which implies that most of the image level features are shared across frames within a video. In such datasets, even a single frame can significantly reduce the uncertainty by providing information about factors of variation shared across frames, such as the person's appearance, background, and even the action being performed. We propose to utilize these properties by modeling the generation process in a hierarchical fashion. Our proposed model first draws a vector encoding a summary frame for the entire video from a fixed prior in the image latent space and subsequently uses an RNN to add variations to it at each time step. The variations over the summary vector are represented as vector additions, which reduce the task of the RNN to performing learned interpolations in the image latent space. Experimental results in both VAE and GAN based models demonstrate that the residual framework ends up mapping the summary vectors to real images, which allows the image generation model to function independently of the computationally expensive RNN. The two-stage generation process, illustrated in Figure~\ref{fig:intro_image}, thus allows performing interpolations in summary frames through a known image latent space to obtain the corresponding variations in videos.

The property of shared content across frames with temporal dependence in the pose can be further exploited by disentangling the underlying factors for video generation into content and motion. A number of prior works achieve this by factorizing each frame's representation into content, and the time-varying component or pose~\cite{MOCOGAN,DSA,MCNET,DRNET}. In this work, we focus on disentanglement into motion and content purely by the design of the probabilistic model without additional loss terms. Two recent approaches, MoCoGAN~\cite{MOCOGAN} and Disentangled Sequential Autoencoder~\cite{DSA} achieved this in the case of VAEs and GANs, respectively. Both the models enforce the content representation to remain fixed across all the frames within a video while the pose or motion vectors vary across frames with temporal dependence. 

We argue that while there is certainly a fixed notion of the objects present in the entire video, it is not reflected exactly in the visual features of individual frames. The temporally dependent pose features govern the variations of different aspects of the appearances of objects to different extents. Thus, a more flexible way to model the disentanglement would be to allow for minor variations in the static attributes even across frames within a video while leaving the temporal dependence to the pose features. We achieve this by using the residual vector technique to encode the content summary for the entire video and subsequently add variations to it at each time step conditioned on the pose vectors. We test the effectiveness of the technique by applying it to MoCoGAN, where it leads to improvement in sample quality while retaining the property of disentanglement. 

\section{Related Work}
A number of prior works have explored the use of deep generative models for the related tasks of video generation and prediction. Unlike unsupervised video generation, the goal of video prediction models is to generate future frames given a number of prior context frames. Both the tasks suffer from high inherent uncertainty and could thus benefit from incorporating knowledge about the properties of video datasets into the model design. 

Ranzato et al.~\cite{VIDEOLANG} adapted language models to video prediction using a dictionary of quantized image patches. Srivastava et al. ~\cite{VIDEOREPR2} used deterministic LSTM encoder-decoder based models for video prediction and unsupervised learning of a fixed-length representation. A number of other approaches focused on deterministic video prediction with a mean squared error loss~\cite{VIDEOREPR2,ATARI}. Mathieu et al.~\cite{MULTISCALE} propose learning strategies to avoid blurry predictions produced by deterministic models trained on mean squared loss, which end up averaging the frames of possible outcomes.

SV2P~\cite{SV2P} proposed a VAE based model for video prediction without a time-varying distribution of latent variables in the generation and inference. SVG~\cite{SVG} improved upon this by using a Variational Recurrent Neural Network based model with a learned prior to flexibly model the dependencies across the latent variables corresponding to the different frames. SAVP~\cite{SAVP} used VAE-GAN hybrid based models to improve the sharpness of the future frame predictions. 

Prior efforts on video generation in the absence of past frames have largely been based on Generative Adversarial Networks. VGAN~\cite{VGAN} used spatio-temporal 3D convolutions to generate videos with a single latent variable to represent the entire video. TGAN~\cite{TGAN} proposed to model the temporal variations by using one-dimensional deconvolutions to generate separate latent variables for individual frames from a single latent variable for the entire video. MoCoGAN~\cite{MOCOGAN} separately encoded the motion and content of a video by dividing each frame's representation into a fixed and a temporally varying part. Disentangled Sequential Autoencoder proposed an analogous model based on VAEs. We further review the details of MoCoGAN and Disentangled Sequential Autoencoder in Section 3. Villegas et al.~\cite{MCNET} separated content and motion representations by using two different encoders, with the motion encoder inferring the representation from the difference of successive frames. Denton et al.~\cite{DRNET} used adversarial and similarity inducing loss terms to disentangle each frame into pose and content. FHVAE\cite{FHVAE} proposed a hierarchical VAE based model for disentangling representations of speech data into sequence-level and segment-level attributes. He et al.~\cite{HOL} used a VAE-based model with structured latent space to allow semi-supervised learning and conditional generation.
Our approach focuses on flexibly sharing information across frames without putting constraints on the image latent space.

\section{Preliminaries: MoCoGAN and Disentangled Sequential Autoencoder}

The architectures of the proposed GAN and VAE based models are based on MoCoGAN and Disentangled Sequential Autoencoder respectively, for a fair comparison with these prior works. By demonstrating the effectiveness of the technique on two families of generative models, we aim to emphasize its general applicability and the resulting interpretation.

Both MoCoGAN and Disentangled Sequential Autoencoder generate a sequence of image representations, one for each frame within the video. Each frame's representation is composed of two random vectors, encoding the content and pose information in the corresponding image separately. The content vector is drawn once from a fixed Isotropic Gaussian prior and remains fixed across all the frames within the video whereas the pose vectors are generated in a sequential manner using a GRU in MoCoGAN and an LSTM in  
Disentangled Sequential Autoencoder. Finally, a CNN based image generator (denoted by $G_I$ in MoCoGAN and our work) is used to map each frame's representation to the corresponding image. In MoCoGAN, an input Gaussian noise is fed at each time step into the GRU to generate the pose vectors having a learnt arbitrary distribution whereas in Disentangled Sequential Autoencoder, the pose vectors are assumed to have an Isotropic Gaussian distribution conditioned on the pose vectors of the previous frames, whose mean and variance are output at each step from the LSTM.

MoCoGAN uses two separate discriminators, $D_V$ and $D_I$, to judge fixed length video clips and individual frames respectively. The video-discriminator $D_V$ is based on spatio-temporal convolutions and evaluates both, the temporal coherence and quality of the video while the image discriminator $D_I$ solely provides feedback to the image generator $G_I$ based on the realism of images. Disentangled Sequential Autoencoder proposed two inference models, named full and factorized, which differed only in the relationship between the content and pose vectors. In the original work, both the models were reported to produce results of similar quality. We thus perform comparisons of our proposed model only with the factorized version. 

\section{Approach}
\begin{figure*}[t]
    \centering
    \begin{minipage}{.45\linewidth}
          \centering
          \centerline{\includegraphics[width=1.0\linewidth]{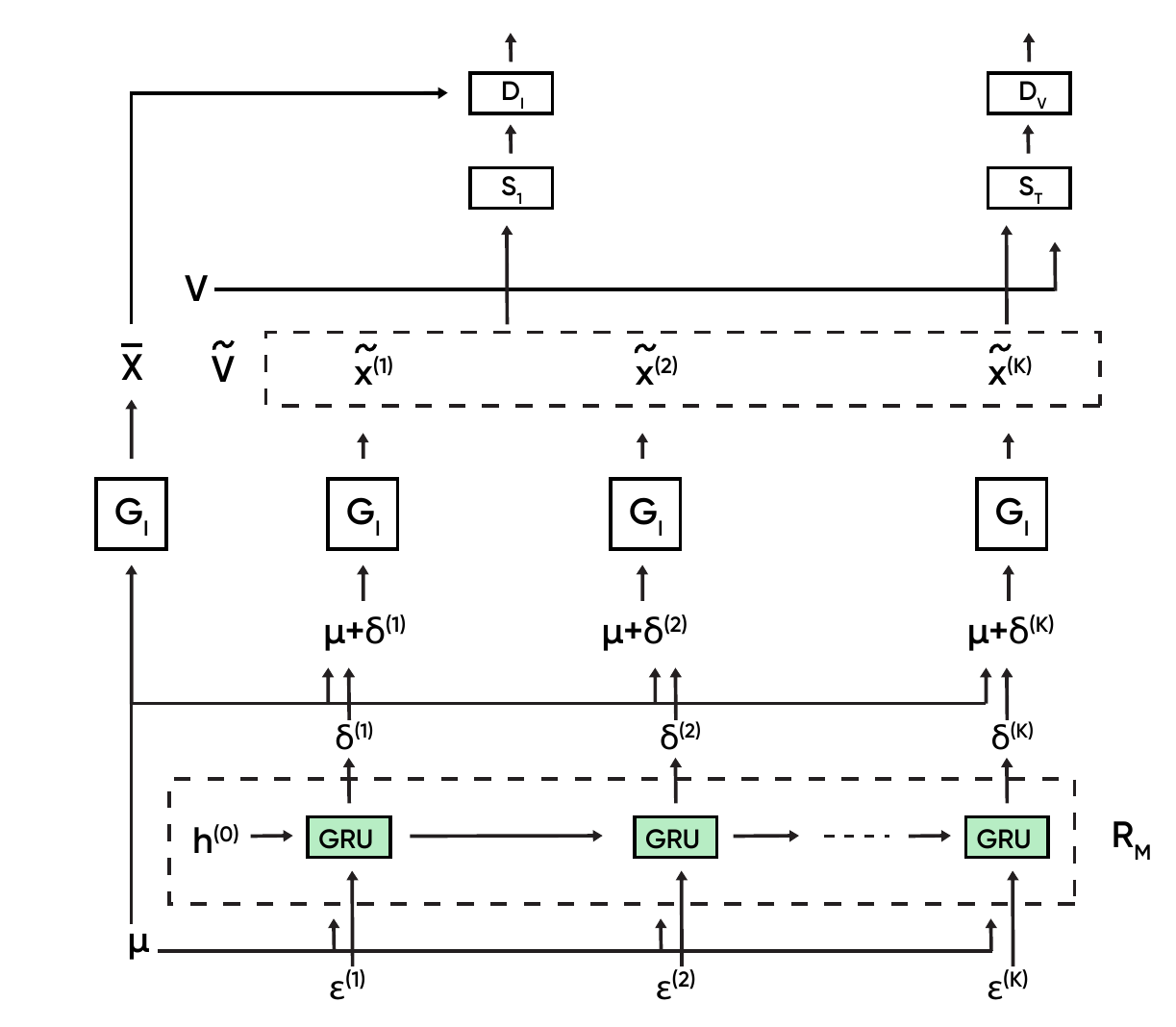}}          
          \centering
          \centerline{(a)}
          \label{fig:GAN_gen}
    \end{minipage}
    \hfill
    \begin{minipage}{.25\linewidth}
          \centering
          \centerline{\includegraphics[width=1.0\linewidth]{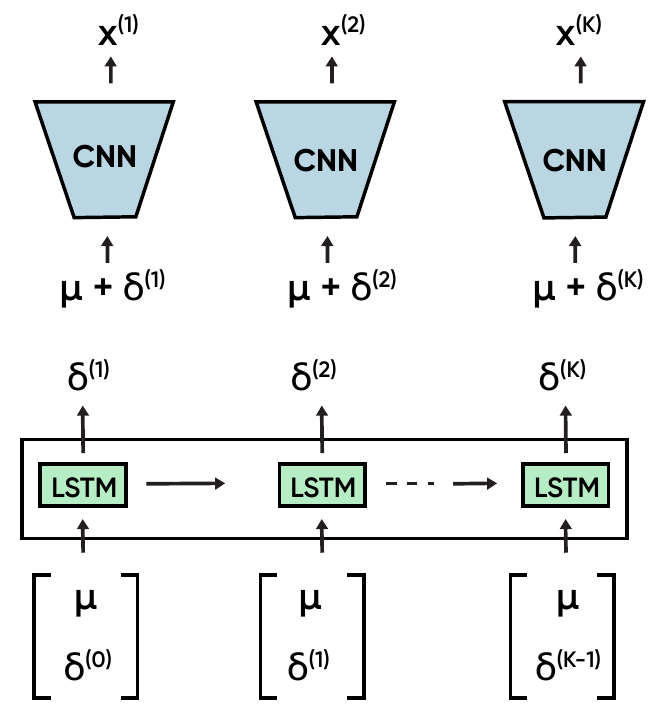}}          
          \centering
          \centerline{(b)}
          \label{fig:VAE_gen}
    \end{minipage}
    \hfill
    \begin{minipage}{.25\linewidth}
       \centering
          \centerline{\includegraphics[width=1.2\linewidth]{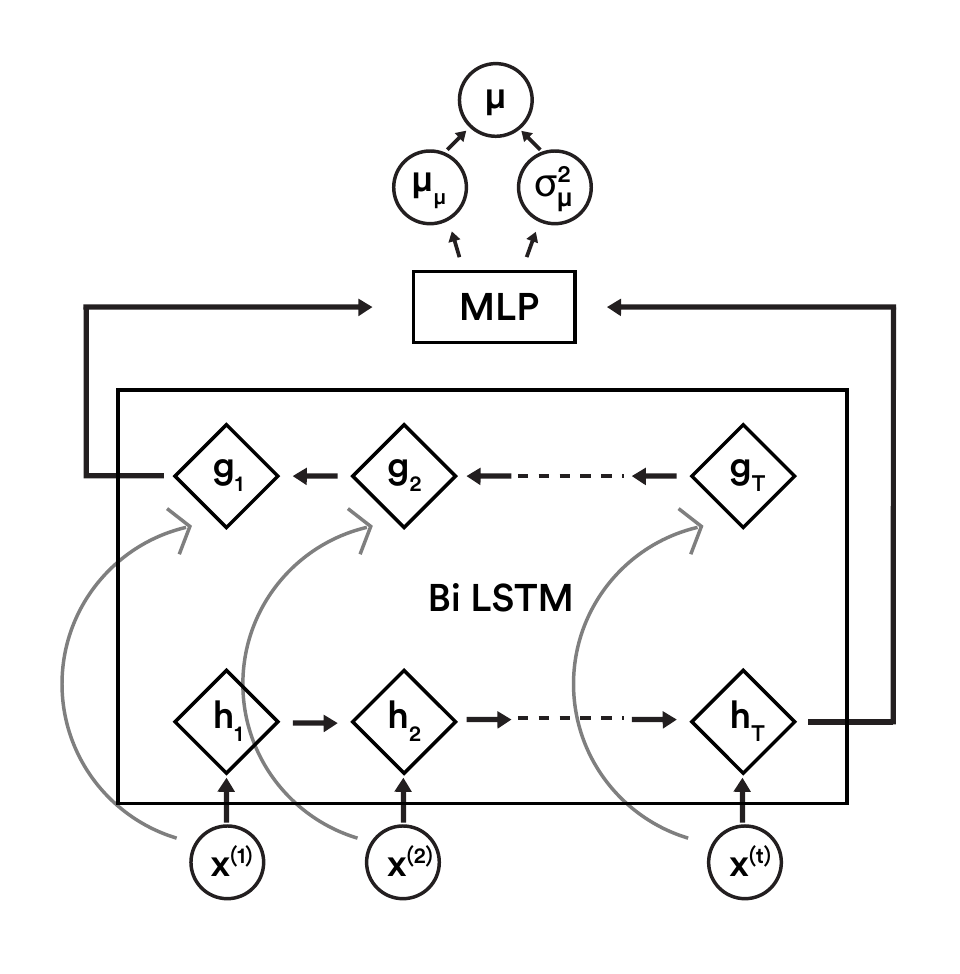}}
          \centering
          \centerline{(c)}
          \centerline{\includegraphics[width=.5\linewidth]{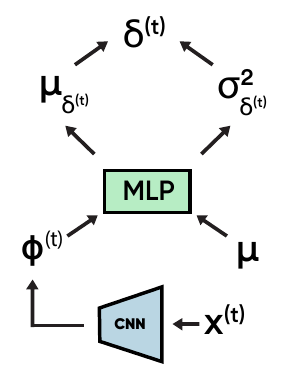}}
          \label{fig:VAE}
          \centerline{(d)}
          \label{fig:VAE_inf1}
    \end{minipage}
    \caption{Illustrations of the proposed summary vector based models: (a) Model structure of RJGAN; (b) Generation of image latent vectors in RJVAE; (c) Inference of summary vector in RJVAE;(d) Inference of image latent vectors in RJVAE}
    \label{fig:arch}
\end{figure*}
    
The central components of the proposed technique are the learning of interpolations in the latent space instead of generation of the entire representation at each step and imposition of an independent fixed prior on the latent vectors. To jointly enforce both the aspects, at each time step, the proposed models add a learned residual vector to a base representation, which we name the summary vector. In section 4.1, we present GAN and VAE video generation models based upon the proposed approach, where the model flexibly exploits similarity across frames without constraining each frame's representation to be disentangled.  In section 4.2, we apply the technique to flexibly achieve disentanglement of each frame's representation into content and motion parts. In both sections, we derive the architecture and notations from the respective baselines, i.e., MoCoGAN and  Disentangled Sequential Autoencoder discussed in section 3.

\subsection{Summary Frame-Based Models}

In the proposed models, named RJGAN and RJVAE for Residual Joint GAN and Residual Joint VAE, respectively, the generation of a video involves first drawing a summary vector in the image latent space, denoted by $\vmu$ and then successively sampling the residual vectors $\{\vd^{(t)}\}_{t=1}^{T}$ conditioned on $\vmu$ as well as the residual vectors of the previous generated frames. The frame level latent variable corresponding to the $t^{th}$ frame, denoted by $\vz^{(t)}$ is then given by $\vmu + \vd^{(t)}$. As illustrated in Figure~\ref{fig:arch} (a) and (b), the temporal dependence across residual vectors in RJGAN and RJVAE is realized in the prior through a GRU (denoted by $R_M$) and an LSTM network respectively following the respective baselines. The prior for  $\vmu$, which is drawn once for each video, is chosen to be Isotropic Gaussian $\mathcal{N}(\vzero, \sigma^2 \vI)$.  At each time step in the generation process, the conditional distribution of the residual vector $p(\vd^{(t)}|\vd^{(<t)},\vmu)$ for RJVAE Figure(~\ref{fig:arch} (b)) is a Multivariate Diagonal Gaussian parameterized by an LSTM cell that receives $\vmu$ as well as the residual vector of the previous frame $\vd^{(t-1)}$ as input while for the RJGAN, Figure(~\ref{fig:arch} (a)) $\vd^{(t)}$ is the output of a GRU cell that receives as input a noise vector $\bm{\varepsilon^{(t)}}$ and $\vmu$. Each frame $\vx^{(t)}$ is generated independently from $\vz^{(t)}$ using a convolutional decoder(denoted by $G_I$ in the RJGAN). The image corresponding to $\vmu$ is denoted by $\bar{\vx}$ and called the summary frame for the video.

As in MoCoGAN, two separate discriminators $D_V$ and $D_I$ are used in RJGAN with inputs as fixed length video clips and individual frames respectively. The inference model for RJVAE, consists of a bi-LSTM network, as shown in Figure~\ref{fig:arch} (c), encoding $q(\vmu | \vx^{(1:T)})$ which learns to extract the factors of variation affecting all the frames and a CNN based encoder, as shown in Figure~\ref{fig:arch} (d), modeling $q(\vd^{(t)}|\vx^{(t)},\vmu)$ which extracts the frame specific variation at each time step.  The full probabilistic model and derivations of the ELBO are provided in the Appendix.

The samples obtained in experiments demonstrate that the summary vectors end up encoding realistic images and exhibit the full range of variations in the various images across videos. Thus, as in image generation models, the learning process automatically ends up imposing the fixed chosen prior on the image representations. The corresponding videos generated by the sequential network end up containing images that share visual features with the summary vector. Moreover, the model also learns to extract information pertaining to temporal factors of variation such as the action being performed from the 
corresponding summary frame.

In MoCoGAN, a video is generated by sampling a length $\kappa$ from a discrete distribution estimated from the data and subsequently running the GRU network $R_M$ for $\kappa$ steps. The minimax objective is then constructed by considering the sampling of random frames and random fixed length clips for $D_I$ and $D_V$ respectively.  The sampling mechanisms for $D_I$ and $D_V$ are represented by functions $S_1$  and $S_T$ respectively which both take as input either a video $\tilde{\vv}$ generated by the model or a real video $\vv$ present in the dataset and output random frames and fixed length clips respectively. The gradient update algorithm updates the set of discriminators $D_I$ and $D_V$, and the combined generator ($G_I$ and $R_M$) in an alternating manner. While following the above training procedure for RJGAN is sufficient to enforce the fixed prior on the image latent space and generate videos of superior quality, we further improve the model's performance by periodically inputting the images corresponding to the summary vector $\vmu$ to the image discriminator $D_I$. Thus we modify the objective function by adding a term corresponding to the discriminator of an image generation model. Denoting real and generated videos by $\vv$ and $\tilde{\vv}$ respectively, the modified objective function is given by:
    \begin{align}
\begin{split}
&\expect_{\vv}[-\log D_I(S_1(\vv))] + \expect_{\tilde{\vv}}[-\log (1-D_I(S_1(\tilde{\vv})))] + \\
    &\expect_{\vv}[-\log D_V(S_T(\vv))] + \expect_{\tilde{\vv}}[-\log (1-D_V(S_T(\tilde{\vv})))] + \\
    &\expect_{\vv}[-\log D_I(S_1(\vv))] + \expect_{\vmu}[-\log (1-D_I(G_I(\vmu))]
\end{split}
\end{align}
    
This modification explicitly enforces the images sampled from the fixed prior to be realistic, following standard image generation models. Evaluation results indicate that this joint sampling of images from videos and the fixed prior leads to improvements in the quality and diversity of both, the generated videos and the sampled images from the fixed prior. 

\subsection{Models with Disentanglement}

The approach based on residual vectors described in the above section can be generalized to any set of vectors with high similarity and shared factors of variation. The major explanatory factors of variation can be captured in a summary vector, while inherent stochastic variations can be represented through residual vectors. In this section, we apply the technique to decompose a video into content and motion without imposing hard constraints of constancy on the image representations. Both MoCoGAN and Disentangled Sequential Autoencoder enforce the content representation to remain constant across the frames within a video. Imposing such a hard constraint of zero variation within a video on certain features of the image representations can hamper the generation capability of the model. For example, in a video of facial expressions, certain visual components such as the appearance of the person's nose can change both within and across videos, albeit to a much different extent. Often, the differentiating factor between the content and pose features is that the pose features are completely responsible for the temporal dependence across frames.

We propose to model the minor variations in the content representation across frames using the residual vector technique. We apply the modification to MoCoGAN to obtain a flexible video generation model capable of disentanglement denoted by RMoCoGAN (Residual MoCoGAN). For the proposed model, we follow the notation and architecture of MoCoGAN. The video generation process, illustrated in Figure\ref{fig:disent} (Appendix), consists of first sampling the content summary vector $\vz_C$ from a fixed Gaussian distribution $\mathcal{N}(\vzero, \sigma^2 \vI)$ and subsequently generating a sequence of motion vectors using the GRU network independent of the content summary. Unlike the baseline, the content representation differs at each time step and is obtained by adding a residual vector $\vd_{C}^{(t)}$ to the content summary. The generation of $\vd_{C}^{(t)}$ is conditioned only on the content summary vector and the motion vector at the $t^{th}$ step. Thus the temporal dependence is captured only through the motion vectors while both, the content and the motion parts of the frames' representations are allowed to vary. Thus the content summary captures the factors of variation independent of temporal dependence and motion while the motion or pose vectors capture the factors affecting the variation over time. Quantitative and qualitative evaluations reveal the improvements obtained by RMoCoGAN in sample quality along with its ability to achieve disentanglement. 
\section{Experiments}
We evaluate the performance of our proposed models on the tasks of video generation (RJGAN and RMoCoGAN), image generation (RJGAN), and the capability of disentangling motion and content (RMoCoGAN). We evaluate our models through human evaluation on 2-AFC (Two Alternative Forced Choice) Tasks, for both image and video generation. For metric based evaluation, we report the Inception Score for image generation and the FVD score for video generation. In addition to the details pertaining to the architecture, hyperparameters, and training conditions of all our models, we provide the specific details of the experiment settings on Amazon Turk, including the question posed, the description provided and the layout in the Appendix.
In order to obtain reliable guarantees of performance improvement, we evaluate our proposed models on three different datasets namely, MUG Facial Expressions, Oulu CASIA NIR \& VIS and Weizmann Action Database. Detailed descriptions of the datasets are provided in the Appendix.

\begin{figure*}[htb]
  \centering
  \includegraphics[width=0.8\linewidth]{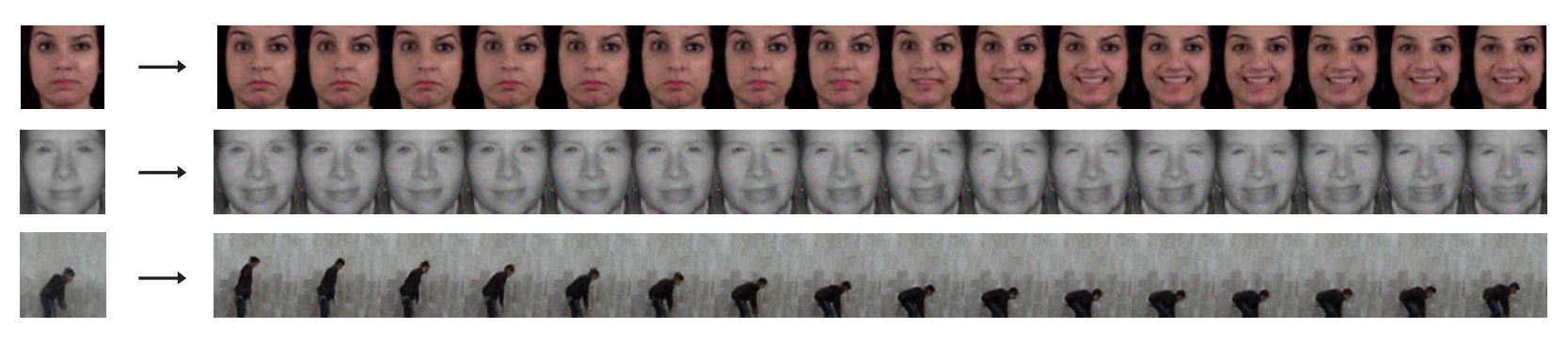}
  \caption[]{Visualization of summary frames and videos generated by RJGAN}
  \label{fig:sumframe}
\end{figure*}

\begin{figure}[htb]
  \centering
  \includegraphics[width=\linewidth]{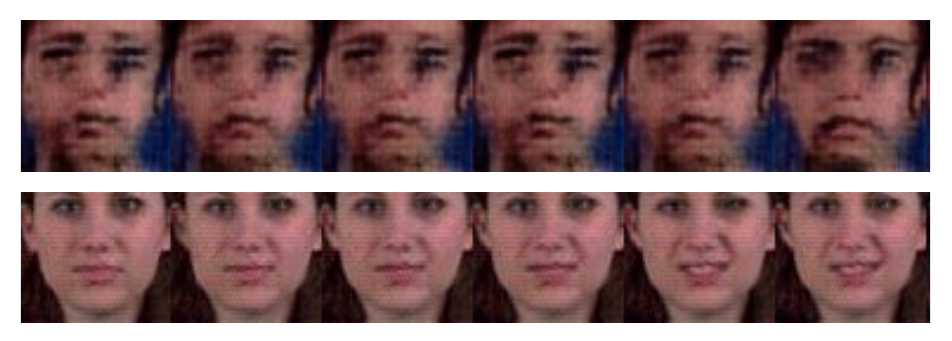}
  \caption[]{Video Generation Samples at 1K iterations from MoCoGAN (top) and RJGAN with Pre-training (bottom) }
  \label{fig:pre}
\end{figure}

\subsection{Video Generation Performance}

\textbf{Human Evaluation} Like ~\cite{MOCOGAN}, we perform human evaluation on 2AFC (Two-alternative Forced Choice) tasks for a quantitative comparison of our approaches with MoCoGAN. For each model under consideration, we randomly generate 128 video sequences for each task and randomly pair the videos for each of our proposed models with that of MoCoGAN, forming 128 2AFC tasks for each pair of models. The videos from two competing models are presented side by side, with the positions randomized in each task to ensure a fair comparison, such that there is an equal probability for a video to occupy the left or right position. The workers are then asked to choose the more realistic video. Each pair of videos is compared by 5 different workers. To ensure the integrity of our results, we selected only the workers that possess a Masters Qualification, obtained by demonstrating continued excellence across a wide range of tasks, and have a lifetime HIT (Human Intelligent Task) approval rate of 95 \%. We report the Mean Worker Preference Scores or the number of times an algorithm is preferred over its competitor, averaged over the number of workers times the number of tasks. As summarised in Table 1, we observe that both RJGAN and RMoCoGAN outperform MoCoGAN on both the datasets. Compared to MoCoGAN, RMoCoGAN achieves a worker preference score of 59.1 \% and 54.7 \% on the MUG and NIR Facial Expressions Dataset respectively, while RJGAN achieves a preference score of 67.2 \% and 57.3 \% respectively. 

\begin{table}[htb]
\begin{center}
\begin{tabular}{|l|c|c|}
\hline
Worker Preference \% & MUG & NIR \\
\hline\hline
MoCoGAN / RMoCoGAN  & 40.9 / \textbf{59.1} & 45.2 / \textbf{54.8} \\
MoCoGAN / RJGAN & 32.8 / \textbf{67.2} & 42.7 / \textbf{57.3}\\
\hline
\end{tabular}
\end{center}
\caption{Worker Preference Scores for Video Generation}
\end{table}

\begin{figure*}[htb]
  \centering
  \includegraphics[width=\linewidth]{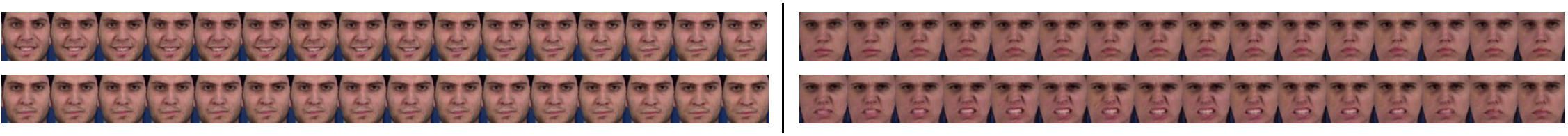}
  \caption[]{Visualization of disentanglement in RMoCoGAN: each half corresponds to a pair of videos generated  by fixing the content vector $\vz_{C}$ and varying the set of motion vectors $\vz_{M}$.}
  \label{fig:disent}
\end{figure*}

\textbf{Metric Based Evaluation} We use the recently proposed metric Frechet Video Distance (FVD) \cite{FVD} to quantitatively evaluate the quality and diversity of the generated videos. A lower score indicates a higher similarity of the distribution of generated videos with the training data distribution. For both datasets, we compute the score by drawing 624 videos each from the real and generated data distributions.

\begin{table}[htb]
\begin{center}
\begin{tabular}{|l|c|c|c|}
\hline
Dataset  & MoCoGAN & RMoCoGAN & RJGAN\\
\hline\hline
MUG &  134.8 & 104.4 & \textbf{99.9} \\
NIR & 125.7 & 118.7& \textbf{105.5} \\
\hline
\end{tabular}
\end{center}
\caption{FVD Scores for Video Generation}
\end{table}

\textbf{Pre-Training and Training on Mixed Datasets} We modify the MUG training dataset by disintegrating half of the videos into individual frames. This results in a dataset consisting of a mixture of images and videos with nearly a $1:1$ split according to the number of images. We verify our model's capability of pre-training in both directions, i.e., training on the set of images first and subsequently training on videos and vice-versa. The latter is verified in Section 5.2 as training the model only on videos leads to improved image generation performance compared to a baseline trained only on images. We test the former quantitatively by comparing the FVD scores after training for 1000 iterations on the set of videos for RJGAN with pre-training, RJGAN without pretraining, and MoCoGAN without pretraining.  Due to the disentanglement constraints, MoCoGAN is incompatible with a pre-trained image generator; as such a generator cannot incorporate constraints specific to videos. Figure~\ref{fig:pre} illustrates qualitative comparison between RJGAN at 1000 iterations after pre-training and MoCoGAN.
\begin{table}[htb]
\begin{center}
\begin{tabular}{|l|c|c|c|}
\hline
& MoCoGAN & RJGAN & RJGAN \\&&&with pre-training\\
\hline\hline
FVD after &&& \\1K iterations &  445.3 & 438.7  & \textbf{199.6} \\
\hline
\end{tabular}
\end{center}
\caption{FVD Scores for Evaluation of Pre-training Capabilities}
\end{table}

We further evaluate RJGAN by training it simultaneously on the 1:1 mixture of videos and images. The Image Discriminator is fed images from both the sets while the Video Discriminator obtains 
only videos. The model achieves an FVD score of $115.6$, surpassing the MoCoGAN model trained on the full set of videos which obtains a score of $134.8$. Unlike MoCoGAN, RJGAN's Image discriminator is not constrained to input images only through the generated videos. The shared enforcement of the fixed prior by the image and video generators allows it to cover the entire distribution of images and account for the missing set of videos.

\textbf{Visualization} We randomly generate 128 videos from RJGAN, and, for each video, visualize the video frames along with the image corresponding to the summary vector, called the summary frame, obtained by providing $\vmu$ as input to $G_{I}$. We observe that this image captures important information shared across the video frames, such as the facial identity in case of the MUG and NIR datasets, and the background, actor identity and action performed in the Weizmann Action dataset. Thus interpolations in image space can be used to obtain the corresponding changes in video space. We also observe that the individual frames of the video depict variations over this summary frame modeled by the additive variations over $\vmu$  in the frame level representations ($\vz^{(t)} = \vmu + \vd^{(t)}$) We present one such illustration each for the MUG, Oulu Casia NIR, and Weizmann Actions datasets in Figure ~\ref{fig:sumframe}, and more examples in the Appendix.

\subsection{Image Generation Performance}
\begin{figure}
  \centering
  \includegraphics[width=\linewidth]{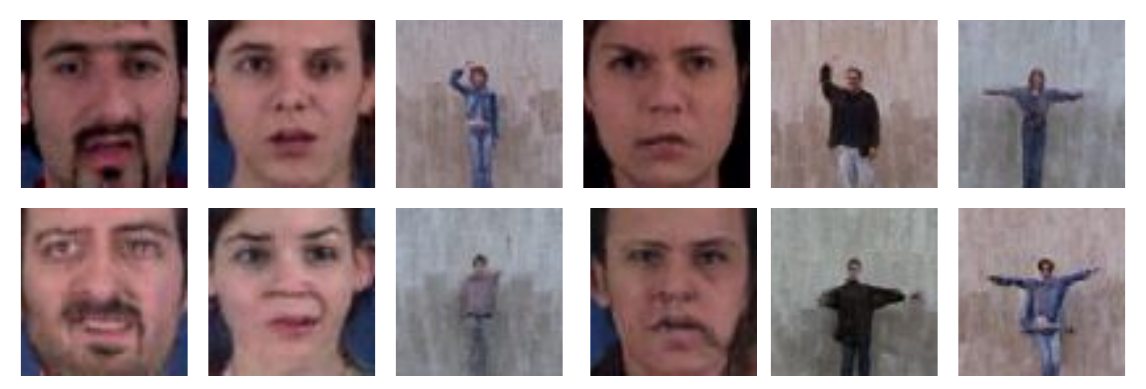}
  \caption[]{Image Generation Samples from RJGAN (top) and Baseline (bottom) }
  \label{fig:imggen}
\end{figure}

We demonstrate the improved image generation capabilities of RJGAN by comparing it with a suitable image generation baseline. The baseline image generator has the same architecture as $G_I$ in RJGAN and is trained to generate individual frames of a video, in an adversarial fashion, using an image discriminator having the same architecture as that of $D_I$ used in RJGAN and MoCoGAN. The image samples from RJGAN are generated by sampling a summary vector $\vmu$ from the Isotropic Gaussian Prior and feeding it to its image generator $G_I$ as input. The proposed technique for image generation is appropriate for benchmarking solely the image generation ability of RJGAN, as it uses only the $G_{I}$ network without interacting with $R_M$ in any manner. For a fair comparison, both the models are trained under the same conditions, details being described in the Appendix. Qualitative and Quantitative results obtained demonstrate that the joint image and video generation paradigm results in improvements in both video generation and image generation tasks, and that supplementing the image generator's learning with temporal information (represented by the additive variations $\vd^{(t)}$) during training time, leads to improved test time performance.

\textbf{Human Evaluation} Similar to the previous section, we perform human evaluation on the image samples generated by our baseline and RJGAN. The complete configuration of the experiment setup, including the reward, workers per task, worker qualifications, and randomization steps, is completely identical to that of the previous sections, except the users are presented two random samples of facial images drawn from RJGAN and the baseline respectively, and asked to choose the more realistic facial image. As summarised in Table 2, we observed that RJGAN outperforms the baseline with a preference score of 64.5 \% on the MUG Facial Expression Dataset. 

\begin{table}[htb]
\begin{center}
\begin{tabular}{|l|c|c|}
\hline
Worker Preference \% & MUG \\
\hline\hline
MoCoGAN / RJGAN & 35.5 / \textbf{64.5} \\
\hline
\end{tabular}
\end{center}
\caption{Worker Preference Scores for Image Generation}
\end{table}

\textbf{Metric Based Evaluation} We train a classifier having the same architecture as the Image Discriminator $D_{I}$ in MoCoGAN and RJGAN, to classify the facial expression from individual, randomly chosen, frames of a video, on the MUG Facial Expression Dataset and use it to calculate the Inception Score as defined in ~\cite{INCEPTION}. We calculate the mean of the Inception Scores averaged over 10 different batches having a batch size of 128, and report the observations and the error estimates in Table 3. We observe that RJGAN obtains a higher Inception Score of 4.50, compared to the baseline, which has an Inception Score of 4.24.

\begin{table}[htb]
\begin{center}
\begin{tabular}{|l|c|c|}
\hline
Dataset  & Baseline & RJGAN \\
\hline\hline
MUG & 4.24 $\pm$ 0.02 & \textbf{4.50 $\pm$ 0.01} \\
\hline
\end{tabular}
\end{center}
\caption{Inception Scores for Image Generation}
\end{table}

\textbf{Visualization} We present some randomly sampled images from RJGAN, on the MUG Facial Expressions and Weizmann Actions datasets in Figure ~\ref{fig:imggen}, and provide more such illustrations in the Appendix. The samples displayed for MUG are a random subset of the summary frames of the videos used during human evaluation.
\subsection{Disentanglement}

\begin{figure}[htb]
  \centering
  \includegraphics[width=\linewidth]{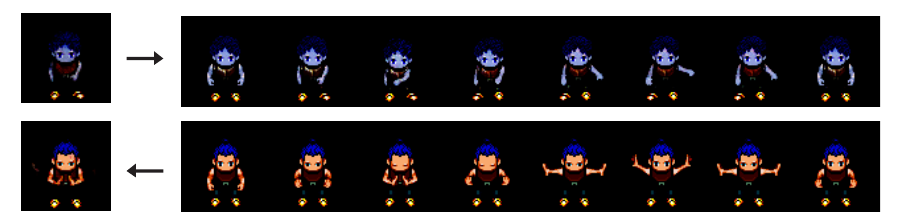}
  \caption[]{Summary frames along with the corresponding videos obtained by generation from the prior (top) and inference from a real video (bottom) in RJVAE}
  \label{fig:vaeframe}
\end{figure}

We perform a qualitative evaluation of the approximate disentanglement of motion and content in RMoCoGAN by first sampling a fixed content representation $\vz_{C}$, and then randomly sampling the sequence of motion vectors $\vz_{M}$ for a fixed number of iterations. In each iteration, a video is generated by first recomputing the sequence of residual vectors $\vd_{C}$ and subsequently generating each frame $x^{(t)}$ by providing $[\vz_{C} + \vd_{C}^{(t)}, \vz_{M}^{(t)}]$ to $G_I$ as input.

Figure ~\ref{fig:disent} illustrates one such example for four different pairs of $\vz_{C}$ and $\vz_{M}$. We observe that the time-invariant aspects such as the facial identity of the person remain constant for a fixed $\vz_{C}$ and varying $\vz_{M}$ only affects the facial expression. The qualitative evaluation of disentanglement, along with the improved Worker Preference Scores of RMoCoGAN over MoCoGAN in Table 1, leads us to conclude that RMoCoGAN improves upon the video generation capabilities of MoCoGAN while still retaining the ability to disentangle content from motion in videos.

\subsection{VAE Based Models}

We evaluate the video generation capabilities of RJVAE and Disentangled Sequential Autoencoder on the Sprites Dataset, which was used for the latter's evaluation in \cite{DSA}. The dataset consists of animated characters with labeled controlled variations in the appearance, in terms of the skin color, tops, pants, and hairstyles (6 varieties for each), as well as the actions, namely walking, spellcasting, and slashing, each present in 3 viewing angles. We use the version of this dataset made publicly available by the authors of \cite{DSA}, and use the same train-test split used by them to train and evaluate the models.

\textbf{Human Evaluation} 
We benchmark RJVAE and the factorized version of Disentangled Sequential Autoencoder by human evaluation on 2-AFC tasks similar to the ones performed in Section 5.1 for MoCoGAN, RMoCoGAN, and RJGAN, except in this case we present the complete frames of the video stacked horizontally, as shown in Figure ~\ref{fig:vaeframe}. This is done due to the small number of frames present in each video and the minute hand movements, which are not clearly visible if an animated video is presented. The procedures followed for the random pairing of the video frames, and the randomization of the position occupied by the videos in the experiment layout is the same as that described in Section 5.1 We provide the workers a short description of the dataset and ask them to choose the video that appears more realistic. We calculate the Mean Worker Preference Scores as defined in Section 5.1 and report them in Table 4 (where Disentangled Sequential Autoencoder is abbreviated as DSA). We find that RJVAE outperforms Disentangled Sequential Autoencoder with a score of 57.7 \% on the Sprites Dataset.  

\begin{table}[htb]
\begin{center}
\begin{tabular}{|l|c|c|}
\hline
Worker Preference \% & Sprites \\
\hline\hline
DSA / RJVAE & 42.3 / \textbf{57.7} \\
\hline
\end{tabular}
\end{center}
\caption{Worker Preference Scores for Video Generation in VAE Models}
\end{table}

\textbf{Visualization} We provide a visualization of both, the generative model and the inference model in RJVAE in Figure 7. The top Figure is a sample from the generative model, generated by drawing $\vmu \sim p(\vmu)$ followed by subsequent sequential sampling of the residual vectors $\vd^{(t)} \sim p(\vd^{(t-1)}, \vmu)$ and the video frames $x^{(1:T)}$, while the bottom figure is a sample from the inference model which draws a random video $x^{(1:T)}$ from the dataset and samples $\vmu$ from the approximate posterior distribution $q(\vmu | x^{(1:T)})$. The summary frame displayed is generated by passing $\vmu$ as input to the CNN generator that maps $\vmu + \vd^{(t)}$ to a video frame $x^{(t)}$. Similar to the visualization in Section 5.1, we observe that the summary frame successfully captures the time-invariant features of the entire video, such as the character identity and clothing, and also indicates the action being performed.

\section{Conclusions and Future Work}
We introduced a novel approach for generating a set of related vectors by encoding the shared information in a summary vector and modeling individual variations as residual vector additions. In the context of video generation, this leads to the design of models wherein image generation is combined with interpolation in the image latent space to generate videos. Experimental evaluations demonstrate improvements in both video and image generation over baselines. Application of the approach to previous models based on disentanglement of video into content and motion leads to increased flexibility in the encoding of images and improved sample quality while retaining the disentanglement capabilities.

Some of the promising directions for future work are (1) applying the proposed technique to datasets involving a mixture of videos and images, (2) pretraining the image generator on large image datasets for subsequent use in videos, (3) generalizing the learned interpolation framework to other transitions in image space such as object rotations and (4) investigating the application of the approach to other domains such as speech and text.

\pagenumbering{roman}
\setcounter{page}{1}
\appendix

\section{Appendix}
\input{supplemental_intro}
\input{experiments_supplemental}

\clearpage

\end{document}

%% file: supplemental_intro.tex
We describe the probabilistic model and derive the ELBO for RJVAE, and provide the details of the generative model for both our GAN based models. We also provide the architecture details and hyperparameter settings of all our models as well as the settings used for the 2-AFC tasks hosted on Amazon Turk. Finally, we offer some visualizations that demonstrate the joint representation learning capabilities of both RJVAE and RJGAN and disentanglement capabilities of RMoCoGAN. 

%% file: experiments_supplemental.tex
\subsection{ELBO Derivation for RJVAE}
\label{sec:app:elborjvae}

The summary vector $\vmu$ is drawn from an Isotropic Gaussian, 
\begin{align*}
p(\vmu) = \Normal(\vmu | \vzero, \sigma_{\vmu}^{2} \textbf{I}).
\end{align*}
The residual vectors $\vd^{(t)}$ are drawn from a Diagonal Gaussian parameterized by an LSTM as follows:

\begin{align*}
p(\vd^{(t)}| \vd^{(<t)}, \vmu) = \Normal(\vd^{(t)} | \mu_{LSTM}(\vmu, \vd^{(t-1)},  h^{(t-1)},c^{(t-1)}), \\ \Sigma_{LSTM}(\vmu, \vd^{(t-1)}, h^{(t-1)}, c^{(t-1)})).
\end{align*}

Where $h^{(t)}$ and $c^{(t)}$ denote the hidden state and cell state of the LSTM at timestep $t$ respectively and $h^{(0)} = c^{(0)} = \vd^{(0)} = 0$. \\
The individual frames of the video are generated as follows:
\begin{align*} p(x^{(t)} | \vmu, \vd^{(t)}) = \Normal(x^{(t)}| G_I(\vmu + \vd^{(t)}), \textbf{I}). 
\end{align*}
Where $G_I$ represents a convolutional image generator that maps each frame level representation to the corresponding video frame.

The complete generative model is as follows:
\begin{align*}
    p(\vmu, \vd^{(1:T)}, x^{(1:T)}) &= p(\vmu) \prod_{t=1}^{T}p(\vd^{(t)}| \vd^{(<t)}, \vmu)p(x^{(t)} | \vmu, \vd^{(t)}).
\end{align*}

The approximate posterior distributions for $\vmu$ and $\vd^{(t)}$ are diagonal Gaussians parameterized by a bi-LSTM and a CNN based encoder with architecture symmetric to $G_I$ respectively.
\begin{align*}
    q(\vmu| x^{(1:T)}) &= \Normal(\vmu | \mu_{biLSTM}(x^{(1:T)}), \Sigma_{biLSTM}(x^{(1:T)}))\\
    q(\vd^{(t)} | \vmu, x^{(t)}) &= \Normal(\vd^{(t)} | \mu_{CNN}(\vmu, x^{(t)}), \Sigma_{CNN}(\vmu, x^{(t)})).
\end{align*}
The inference model is as follows \\
$q(\vmu, \vd^{(1:T)} | x^{(1:T)}) = q(\vmu| x^{(1:T)})\prod_{t=1}^{T}q(\vd^{(t)} | \vmu, x^{(t)})$. \\

The model is trained by maximising the ELBO $\mathcal{L}$, 
\begin{align*}
    & \mathcal{L}  = 
    \expect_{q(\vmu, \vd^{(1:T)} | x^{(1:T)})}[\displaystyle \log \cfrac{p(\vmu, \vd^{(1:T)}, x^{(1:T)})}{q(\vmu, \vd^{(1:T)} | x^{(1:T)})}]\\
    &= \sum_{t=1}^{T} \expect_{q(\vmu| x^{(1:T)})q(\vd^{(t)} | \vmu, x^{(t)})}[\log p(x^{(t)} | \vmu, \vd^{(t)})] \\
    & - \sum_{t=1}^{T} \expect_{q(\vmu| x^{(1:T)})}[D_{KL}[q(\vd^{(t)} | \vmu, x^{(t)}) || p(\vd^{(t)}| \vd^{(<t)}, \vmu)]] \\
    & - D_{KL}[q(\vmu| x^{(1:T)}) || p(\vmu)].
\end{align*}

\subsection{Generative Model for RJGAN}
\label{sec:app:genmodelrjgan}
The summary vector $\vmu$ is drawn from the Isotropic Gaussian, $p(\vmu) = \Normal(\vmu | 0, \sigma_{\vmu}^{2} \textbf{I})$. The residual vectors $\vd^{(t)}$ are generated by a nonlinear transformation (represented by the GRU network $R_M$) of $[\vmu, \epsilon^{(t)}]$ with $\epsilon^{(1)}, \epsilon^{(2)}, \cdots, \epsilon^{(T)} \stackrel{iid}{\sim} \Normal(\epsilon| 0, \sigma_{\epsilon}^{2}\mathbf{I})$. The initial hidden state of the GRU is given as $h^{(0)} = 0$. \\
The individual frames of the video are generated by the convolutional image generator $G_I$ as follows:
$$\Tilde{x^{(t)}} = G_I(\vmu + \vd^{(t)}).$$
The model is trained using the adversarial objective function ($1$) defined in Section 4.1 

\subsection{Generative Model for RMoCoGAN}
\label{sec:app:genmodelrjgan}
The content code $\vz_C$ is drawn from the Isotropic Gaussian, 
$p(\vz_C) = \Normal(\vz_C | \vzero, \sigma_{\vz_C}^{2} \textbf{I})$. The motion vectors $\vz_{M}^{(t)}$ are generated by a nonlinear transformation (represented by the GRU network $R_M$) of $\epsilon^{(t)}$ with $\epsilon^{(1)}, \epsilon^{(2)},\cdots, \epsilon^{(T)} \stackrel{iid}{\sim} \Normal(\epsilon| 0, \sigma_{\epsilon}^{2}\mathbf{I})$. The initial hidden state of the GRU is given as $h^{(0)} = 0$. \\ The residual content vectors $\vd_{C}^{(t)}$ are generated by a non-linear transformation of $[\vz_C, \vz_{M}^{(t)}]$ represented by a two hidden layer MLP $NN_{\vd}$ such that
$\vd_{C}^{(t)} = NN_{\vd}(\vz_C, \vz_{M}^{(t)})$. \\ 
The individual frames of the video are generated by the convolutional image generator $G_I$ as follows: \\
$$\Tilde{x}^{(t)} = G_I(\vz_C + \vd_{C}^{(t)}, \vz_{M}^{(t)}).$$\\
The model is trained using the same objective as MoCoGAN. 

\begin{figure}
  \centering
  \includegraphics[width=\linewidth]{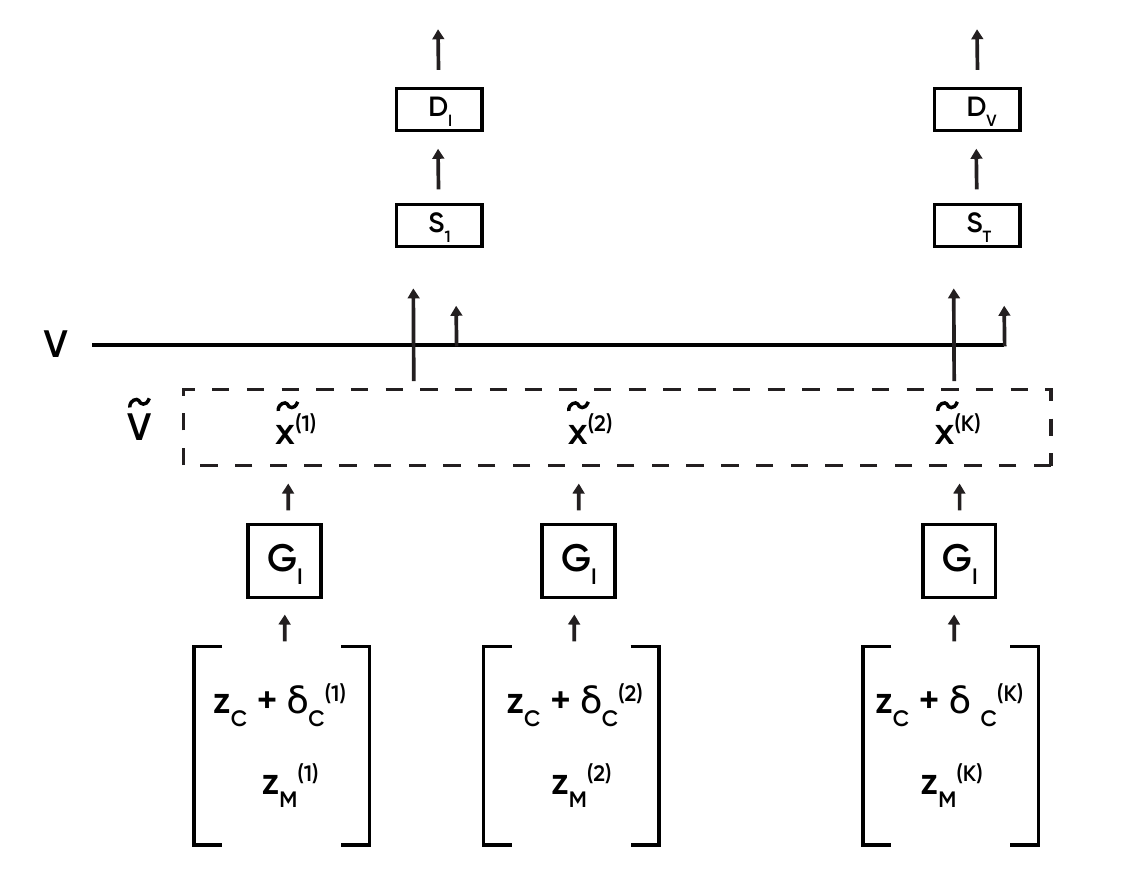}
  \includegraphics[width=\linewidth]{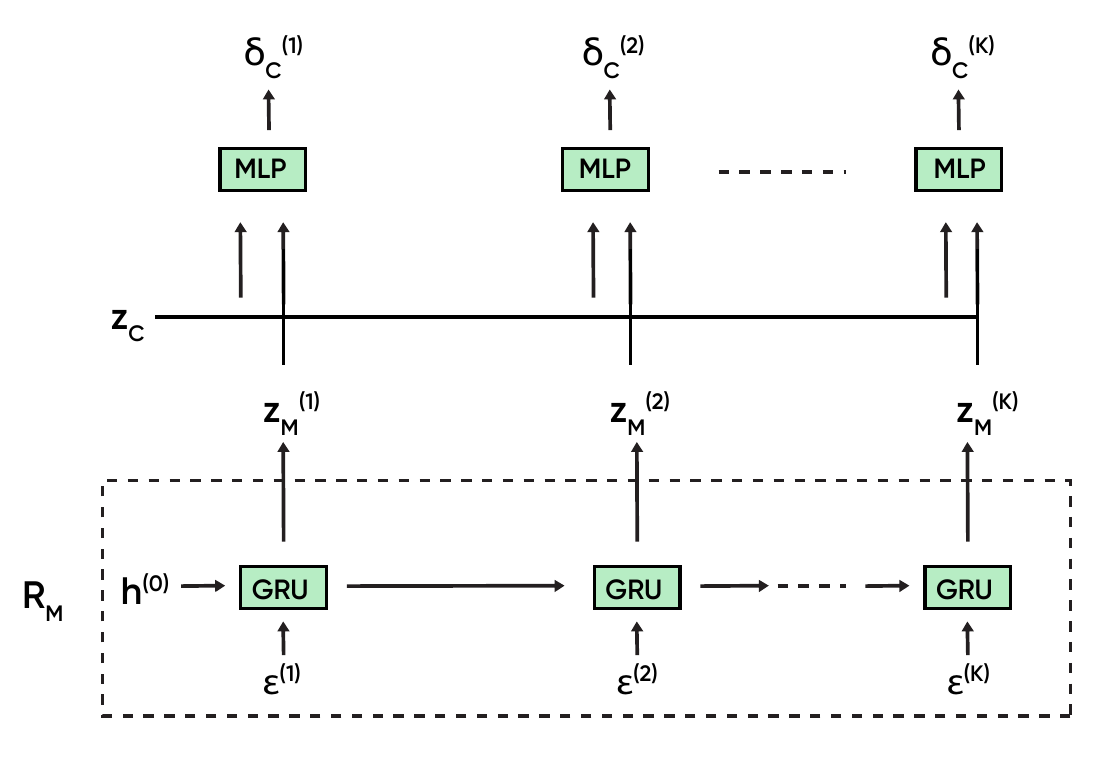}
  \caption[]{Model structure of RMoCoGAN}
  \label{fig:disent}
\end{figure}

\subsection{Datasets}
We evaluate our GAN based video generation models on the following datasets:
\begin{itemize}
    \item \textbf{MUG Facial Expressions} The MUG Facial Expressions Database ~\cite{MUG}  contains 86 subjects performing facial expressions, out of which only 52 are available to authorized online users. We trained our proposed GAN models and benchmarked them against MoCoGAN on the subset of 52 subjects available online. Similar to the pre-processing employed in ~\cite{MOCOGAN}, we considered only the videos that are at least 64 frames long and belong to one of the following facial expression categories: anger, disgust, happiness, sadness, and surprise. We used the pretrained HOG Face Detector available in dlib ~\cite{DLIB} to detect and crop the facial regions from each frame, and scaled them to $64 \times 64$. In the end, our training set comprised of $633$ videos.
    
    \item \textbf{Oulu CASIA NIR \& VIS} The Oulu-CASIA NIR \& VIS Facial Expression Database ~\cite{CASIA}  contains videos from 80 subjects performing the following facial expressions: anger, disgust, happiness, sadness, and surprise. The videos are captured with two imaging systems, NIR (Near Infrared) and VIS (Visible light), under three different illumination conditions for each: normal indoor illumination, weak illumination, and dark illumination. We use the NIR videos as our training data after applying the same pre-processing and filtering as the MUG dataset.
    
    \item \textbf{Weizmann Action Database} We use the pre-processed version of the Weizmann Action Database ~\cite{WEIZMANN} bundled within the official implementation of MoCoGAN, which consists of 72 videos of 9 people performing the following actions in various backgrounds: one-handed wave, two-handed wave, jumping jack and bend, with the video frames scaled to $64 \times 64$. Similar to MoCoGAN, we only perform a qualitative evaluation on this dataset, due to its small size, and provide a visualization in Figure ~\ref{fig:sumframe} and Figure ~\ref{fig:imggen}
\end{itemize}
\subsection{Architecture and Hyperparameters}
\label{sec:app:architecture}

\paragraph{RJGAN} 
For fair comparison, the architecture of the image generator $G_I$, GRU $R_M$, and the image and video discriminators $D_I$ and $D_V$ are the same as the ones used by MoCoGAN, with changes only in the type and dimension of input where required. The latent code $\vz^{(t)} = \vmu + \vd^{(t)}$ input to $G_I$ is of dimension 60, which is the same as the dimension of the image latent space in MoCoGAN. The input to $R_M$ is at each timestep concatenation of $\vmu$ and $\epsilon^{(t)}$ where the dimension of $\epsilon^{(t)}$ is also 60. Both the variance parameters $\sigma_{\vmu}$ and $\sigma_{\epsilon}$ are set to 1. The model is trained under the same training conditions as MoCoGAN, using the Adam Optimizer with a learning rate of 0.0002, $\beta_1$ and $\beta_2$ set to 0.5 and 0.999 respectively with a weight decay of $10^{-5}$, trained for 100,000 iterations with a batch size of 32. 

\paragraph{RMoCoGAN}

As in RJGAN, for fair comparison, the architecture of the different components and the dimensions of the content code $\vz_C$, motion code $\vz_M$ and the noise $\epsilon$ are the same as the ones used by MoCoGAN. The nonlinear mapping from $[\vz_C, \vz_{M}^{(t)}]$ to $\vd_{C}^{(t)}$ denoted by $NN_{\vd}$ is a two hidden layer MLP. The hidden layers comprise an affine transform, 1-D Batch Normalization and ReLU nonlinearity, while the output layer is simply an affine transform. The first hidden layer maps the 60 dimensional input $[\vz_C, \vz_{M}^{(t)}]$ to a 50 dimensional representation while the subsequent layers do not change the dimension.  Training conditions, batch size and optimizer settings are the same as MoCoGAN and RJGAN.

\paragraph{RJVAE}
As the source code of Disentangled Sequential Autoencoder was not made publicly available by the authors, we attempted to reproduce the model, strictly adhering to the architecture details and hyperparameters provided. The hyperparameters not provided in the paper were the optimizer configuration, batch size and number of epochs trained. For both RJVAE and the baseline, we used the Adam Optimizer with a learning rate of 0.0005, $\beta_1$ and $\beta_2$ set to 0.9 and 0.999 respectively without any weight decay, and trained for 600 epochs with a batch size of 128. The image encoder $G_I$ for both models have the same architecture. The prior LSTM and the posterior inference network for the residual vectors $\vd^{(t)}$ in RJVAE have the same architecture as those of the dynamics encodings $z_t$ in Disentangled Sequential Autoencoder, and the same is true for the bi-LSTM inference network for the summary vector $\vmu$ in RJVAE and the content encoding $f$ in Disentangled Sequential Autoencoder. For RJVAE, the latent code $\vz^{(t)} = \vmu + \vd^{(t)}$ input to $G_I$ is of dimension 64. At each timestep, the input to the prior of $\vd^{(t)}$ is the concatenation of $\vmu$ and $\vd^{(t-1)}$ each having dimension 64, and the input to the posterior inference network for $\vz^{(t)}$ is a concatenation of $\vphi^{(t)}$ and $\vmu$ having dimension 512 and 64 respectively, where $\vphi^{(t)}$ is the encoding of the frame $x^{(t)}$ generated by the CNN encoder.   

\subsection{An Alternative to the Summary Frame Based Model}

An alternative approach to applying the residual vector framework for video generation is to generate at a time step, the vector difference between the current and the previous frame. Thus we would have $\vz^{(t)} = \vd^{(t)} + \vz^{(t-1)}$. In such a model, sampling from a fixed prior can be used for generation of the first frame's representation $\vz^{(1)}$ or an imagined zeroth frame's representation $\vz^{(0)}$. We experimentally validated our proposed summary vector based model's superiority over the above approach. Firstly, the above approach doesn't learn to impose the fixed prior over the entire set of images in datasets where any arbitrary image can't be the first frame of a video, leading to image generation of greatly reduced diversity. Secondly, the first/zeroth frame approach can't encode video level factors of variation in a shared vector that affects all the frames. This lack of hierarchical structure leads to reduced temporal coherence in the generated videos. In the proposed summary frame based models, $\vmu$ can encode any arbitrary frame and provides a channel to carry information across the entire video.
\subsection{Amazon Turk Settings}
\label{sec:app:amt}
The worker qualifications, workers per task and the randomization procedures employed for each of the human evaluation tasks have already been discussed in the main paper. In this section, we describe the other factors involved such as the title and description for each task, the time limit and the reward involved.  

\paragraph{Video Generation for MUG and Oulu CASIA NIR}
The title of each 2-AFC task was  ``Pick the facial expression video that looks more realistic", accompanied by the following description.  ``You are presented with a pair of videos of human facial expressions, both generated from deep learning models trained on real videos. Each video clip is less than a second long. You are requested to choose the video clip that appears more realistic to you". A time limit of two minutes was assigned to each task and each worker received \$ 0.03 for completing a task.

\paragraph{Video Generation for Sprites}
Each task was titled ``Which of the following videos of animated characters appear more realistic, considering the dataset description provided?" and the description provided was: ``The dataset contains animated characters performing actions like walk, spellcast and slash from three viewing angles. For a realistic video, the action performed is identifiable and the character's appearance doesn't change.". A time limit of two minutes was assigned to each task and each worker received \$ 0.03 for completing a task. 

\paragraph{Image Generation for MUG}
Each task was titled  ``Pick the facial image that looks more realistic" and contained the description  ``You are presented with a pair of images of human faces, both generated from deep learning models. Choose the facial image that appears more realistic to you.". A time limit of one minute was assigned to each task and each worker received \$ 0.03 for completing a task.

\subsection{Visualization}
\label{sec:app:qualitativeanalysis}
We provide illustrative examples of disentanglement of motion and content in RMoCoGAN in Figure 8, by showing three different facial expressions each for two facial identities, generated by fixing the content vector $\vz_C$ for each facial identity whilst varying the motion vectors $z_M$. Figure 9 visualizes the latent space of the image generator $G_I$ for both RJVAE and RJGAN by sampling two distinct summary frame representations, interpolating between them, and displaying the corresponding summary frames. Figures 10, 11 and 12 visualize the joint image and video generation capabilities by interpolating between two summary frame representations, and using the interpolated representation to generate both the summary frames and the corresponding videos.

\begin{figure*}[htb]
  \centering
  \includegraphics[width=\linewidth]{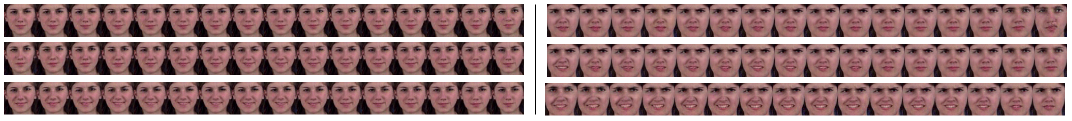}
  \caption[]{Visualization of disentanglement in RMoCoGAN: each half corresponds to a set of three videos generated by fixing the content vector $\vz_{C}$ and varying the set of motion vectors $\vz_{M}$.}
  \label{fig:disent}
\end{figure*}

\begin{figure*}[htb]
  \centering
  \includegraphics[width=\linewidth]{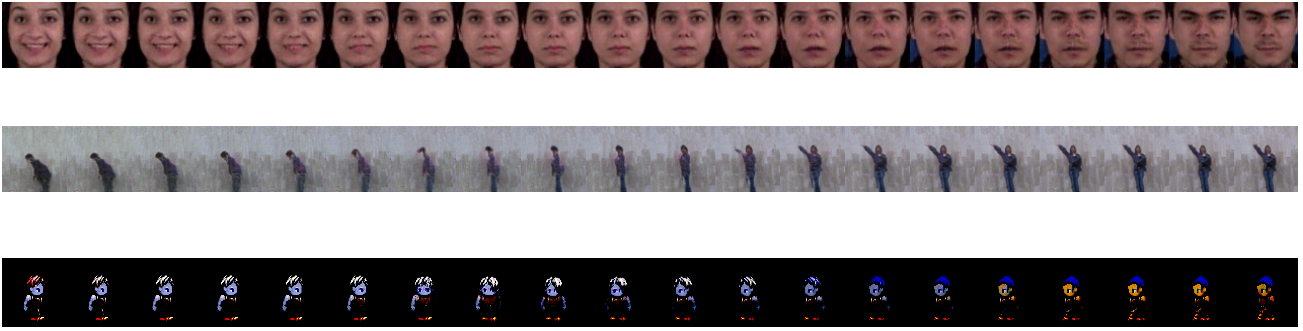}
  \caption[]{Image Space Interpolation in RJGAN on the MUG Facial Expressions (Top) and Weizmann Actions Datasets (Middle), and in RJVAE on the Sprites Dataset (Bottom).}
  \label{fig:disent}
\end{figure*}

\begin{figure*}[htb]
  \centering
  \includegraphics[width=\linewidth]{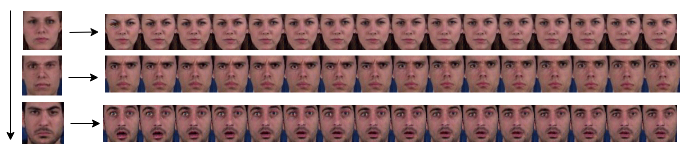}
  \caption[]{Joint image and video generation in RJGAN on the MUG dataset: interpolation in videos can be obtained by interpolation of the summary frames in image space.}
  \label{fig:disent}
\end{figure*}

\begin{figure*}[htb]
  \centering
  \includegraphics[width=\linewidth]{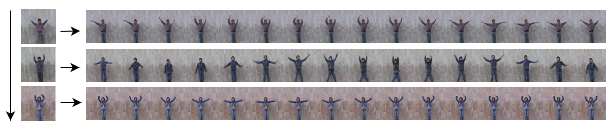}
  \caption[]{Joint image and video generation in RJGAN on the Weizmann dataset: interpolation in videos can be obtained by interpolation of the summary frames in image space.}
  \label{fig:disent}
\end{figure*}

\begin{figure*}[htb]
  \centering
  \includegraphics[width=\linewidth]{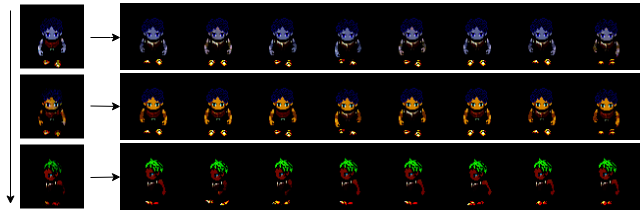}
  \caption[]{Joint image and video generation in RJVAE on the Sprites dataset: interpolation in videos can be obtained by interpolation of the summary frames in image space.}
  \label{fig:disent}
\end{figure*}